\documentclass[10pt,twocolumn,letterpaper]{article}

\usepackage{cvpr}
\usepackage{times}
\usepackage{epsfig}
\usepackage{graphicx}
\usepackage{amsmath}
\usepackage{amssymb}
\usepackage{booktabs}
\usepackage{arydshln}
\usepackage{multirow}
\usepackage{latexsym}
\usepackage{graphicx}
\usepackage{comment}
\usepackage{enumitem}
\usepackage{enumerate}
\usepackage{paralist}
\usepackage{xcolor}

 \renewenvironment{itemize}[1]{\begin{compactitem}#1}{\end{compactitem}}

\usepackage[pagebackref=true,breaklinks=true,letterpaper=true,colorlinks,bookmarks=false]{hyperref}

 \cvprfinalcopy 

\def\cvprPaperID{10586} 

\ifcvprfinal\fi
\begin{document}

\title{Multimodal Categorization of Crisis Events in Social Media}

\author{Mahdi Abavisani\thanks{Equal contribution, with ordering decided by Python. Research work was done while authors
were interning at Dataminr Inc.}\\
Dataminr Inc., New York, NY\\
{\tt\small mabavisani@dataminr.com}
\and
Liwei Wu$^*$\\
Department of Statistics\\
University of California, Davis\\
Davis, CA\\
{\tt\small liwu@ucdavis.edu}
\and
Shengli Hu\\
Dataminr Inc., New York, NY\\
{\tt\small shu@dataminr.com}
\and
Joel Tetreault\\
Dataminr Inc., New York, NY\\
{\tt\small jtetreault@dataminr.com}
\and
Alejandro Jaimes\\
Dataminr Inc., New York, NY\\
{\tt\small ajaimes@dataminr.com}
}

\maketitle

\begin{abstract}
Recent developments in image classification and natural language processing, coupled with the rapid growth in social media usage, have enabled fundamental advances in detecting breaking events around the world in real-time.  Emergency response is one such area
that stands to gain from these advances.  By processing billions of texts and images a minute, events can be automatically detected to enable emergency response workers to better assess rapidly evolving situations and deploy resources accordingly.  To date, most event detection techniques in this area have focused on image-only or text-only approaches, limiting detection performance and impacting the quality of information delivered to crisis response teams.  In this paper, we present a new multimodal fusion method that leverages both images and texts as input.  In particular, we introduce a cross-attention module that can filter uninformative and misleading components from weak modalities on a sample by sample basis. In addition, we employ a multimodal graph-based approach to stochastically transition between embeddings of different multimodal pairs during training to better regularize the learning process as well as dealing with limited training data by constructing new matched pairs from different samples. We show that our method outperforms the unimodal approaches and strong multimodal baselines by a large margin on three crisis-related tasks.

\end{abstract}

\section{Introduction}
Each second, billions of images and texts that capture a wide range of events happening around us are uploaded to social media platforms from all over the world.  At the same time, the fields of Computer Vision (CV) and Natural Language Processing (NLP) are rapidly advancing \cite{howard2017mobilenets,he2017mask,devlin2018BERT} and are being deployed at scale. With large-scale visual recognition and textual understanding available as fundamental tools, it is now possible to identify and classify events across the world in real-time. This is possible, to some extent, in images and text separately, and in limited cases, using a combination. A major difficulty in crisis events,\footnote{An event that is going (or is expected) to lead to an unstable and dangerous situation affecting an individual, group, community, or whole society (from Wikipedia); typically requiring an emergency response.} in particular, is that as events surface and evolve, users post fragmented, sometimes conflicting information in the form of image-text pairs. This makes the automatic identification of notable events significantly more challenging. 

\begin{figure}[t]
    \centering
    \includegraphics[width=0.35\textwidth]{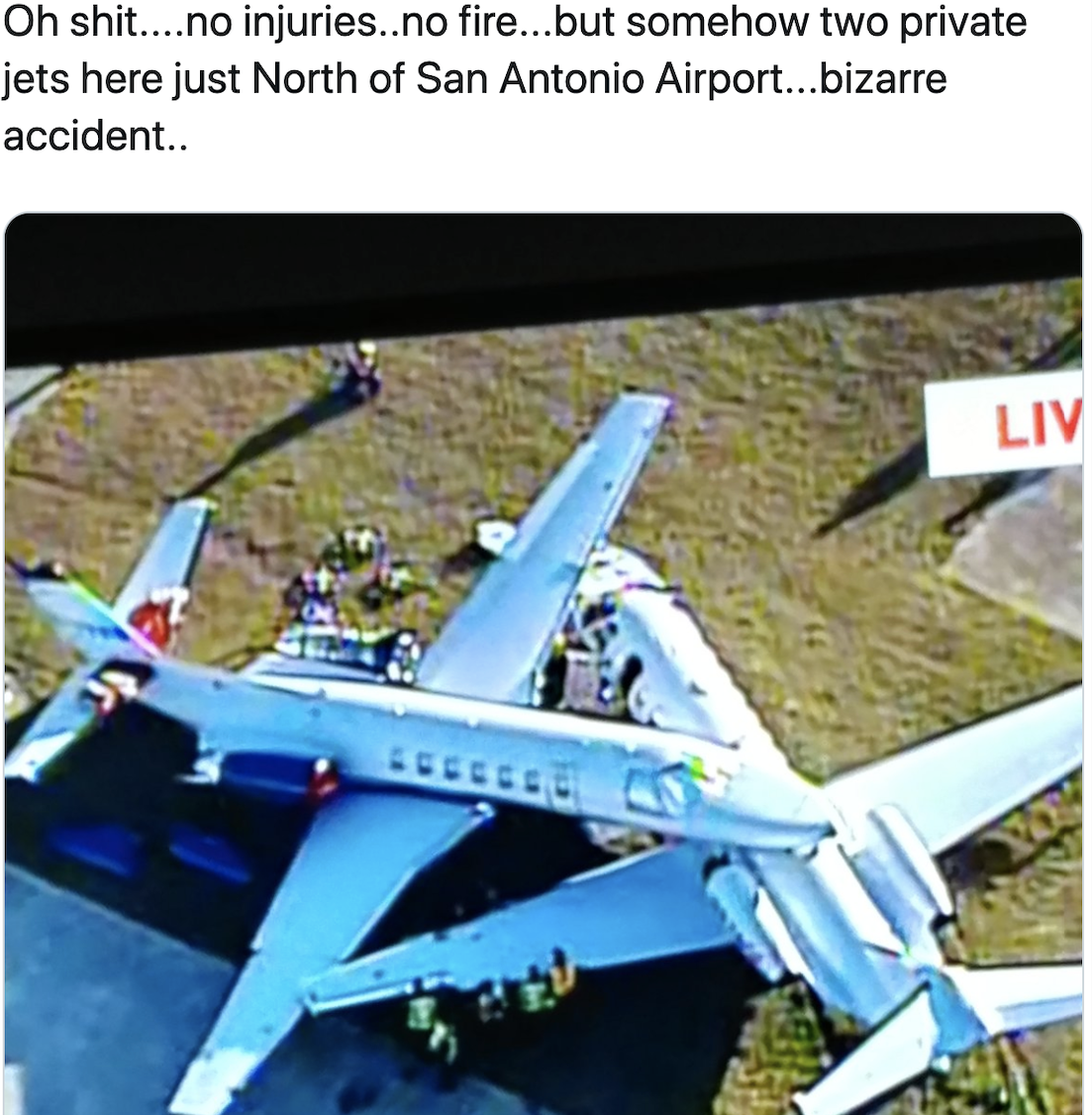}
    \label{fig:txtimg}
    \caption{A Crisis-related Image-text Pair from Social Media}
\end{figure}

Unfortunately, in the middle of a crisis, the information that is valuable for first responders and the general public often comes in the form of image-text pairs.  So while traditional CV and NLP methods that treat visual and textual information separately can help, a big gap exists in current approaches. Despite the general consensus on the importance of using AI for Social Good \cite{harding2015breast,gebru2017using,abebe2019using}, 
the power of social media, and a long history of interdisciplinary research on humanitarian crisis efforts, there has been very little work on automatically detecting crisis events {\em jointly} using visual and textual information.  

Prior approaches that tackle the detection of crisis events have focused on either image-only or text-only approaches.  
As shown in Figure~\ref{fig:txtimg}, however, an image alone can be ambiguous in terms of its urgency whereas the text alone may lack details. 

\begin{figure}[t!]
    \centering
    \includegraphics[width=0.48\textwidth]{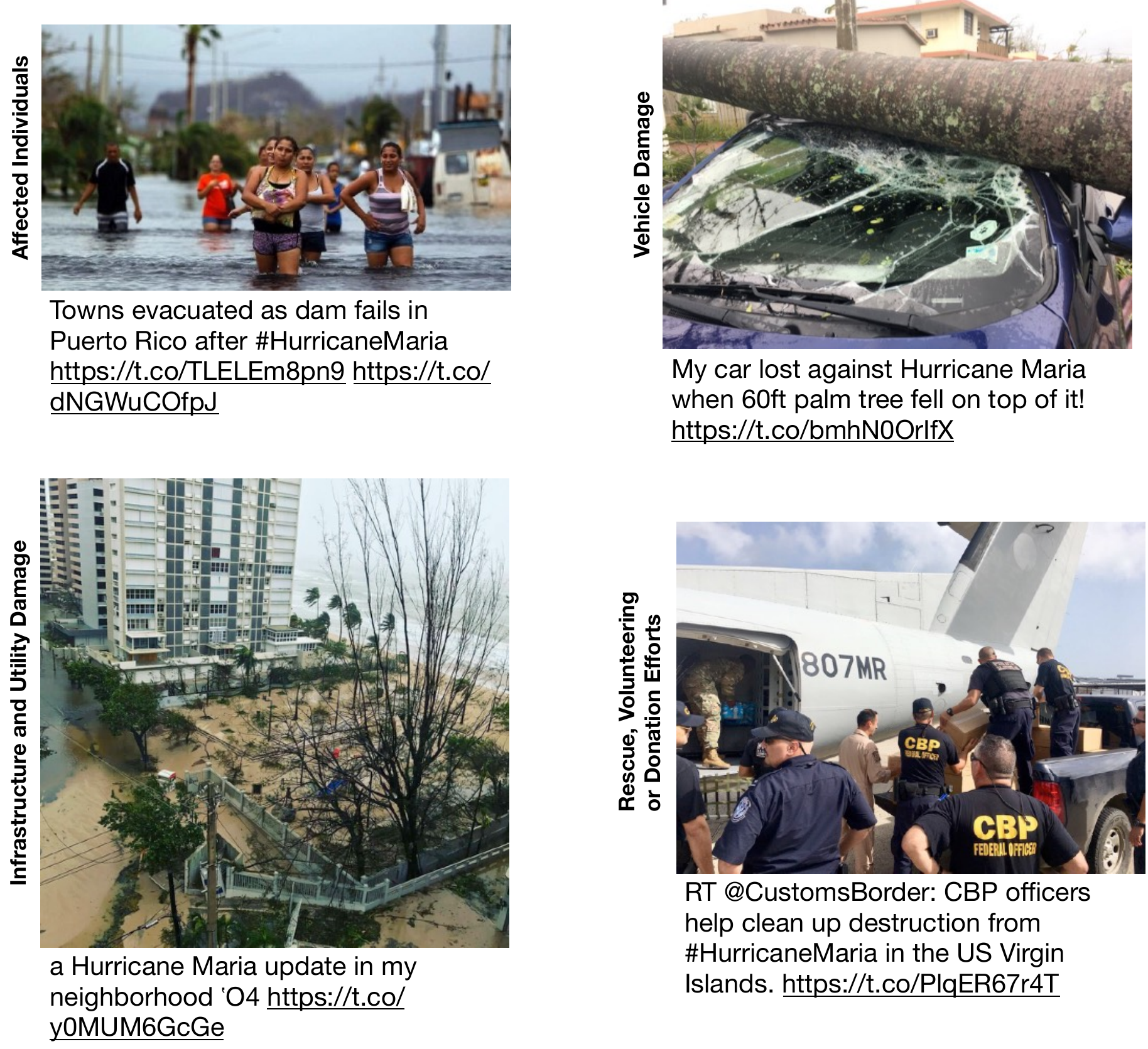}
    \caption{Samples from Task 2; Event Classification with Texts and Images.}\label{fig:classes}
\end{figure}

To address these issues, we propose a framework to detect crisis events using a combination of image and text information.  In particular, we present an approach to automatically label images, text, and image-text pairs based on the following criteria/tasks:  1) \textbf{Informativeness}: whether the social media post is useful for providing humanitarian aid in an emergency event, 2) \textbf{Event Classification}: identifying the type of emergency (in Figure~\ref{fig:classes}, we show some of the categories that different image-text pairs belong to in our event classification task), and 3) \textbf{Severity}: rating how severe the emergency is based on the damage indicated in the image and text. Our framework consists of several steps in which, given an image-text pair, we create a feature map for the image, generate word embeddings for the text, and propose a cross-attention mechanism to fuse information from the two modalities. It differs from previous multimodal classification in how it deals with fusing that information. 

In short, we present a novel, multimodal framework for classification of multimodal data in the crisis domain.  This approach, "Cross Attention", avoids transferring negative knowledge between modalities and makes use of stochastic shared embeddings to mitigate overfitting in small data as well as dealing with training data with inconsistent labels for different modalities. Our model outperforms strong unimodal and multimodal baselines by up to 3 F-score points across three crisis tasks.

%
%
%
%

\section{Related Work}
\noindent\textbf{AI for Emergency Response:}
Recent years have seen an explosion in the use of Artificial Intelligence for Social Good \cite{harding2015breast,gebru2017using,abebe2019using}. 
Social media has proven to be one of most relevant and diverse resources and testbeds, whether it be for identifying risky mental states of users \cite{buechel2018modeling,eichstaedt2018facebook,guntuku2019twitter}, recognizing emergent health hazards \cite{eichstaedt2018more}, filtering for and detecting natural disasters \cite{said2019natural,madichetty2019detecting,rudnerrapid}, or surfacing violence and aggression in social media \cite{blevins-etal-2016-automatically}. 

Most prior work on detecting crisis events in social media has focused on text signals. 
For instance, Kumar \etal~\cite{kumar2011tweettracker} propose a real-time tweet-tracking system to help first responders gain situational awareness once a disaster happens. Shekhar \etal~\cite{shekhar2015disaster} introduce a crisis analysis system to estimate the damage level of properties and the distress level of victims.  At a large scale, filtering (e.g., by anomaly or burst detection), identifying (e.g., by clustering), and categorizing (e.g., by classifying) disaster-related texts on social media have been the foci of multiple research groups \cite{stowe2016identifying,to2017identifying,yin2015using}, achieving accuracy levels topping at $0.75$ on small annotated datasets collected from Twitter.  

Disaster detection in images has been an active front, whether it be user-generated content or satellite images (for a detailed survey, refer to Said \etal~\cite{said2019natural}). For instance, Ahmad \etal~\cite{ahmad2017jord} introduce a pipeline method to effectively link remote sensor data with social media to better assess damage and obtain detailed information about a disaster. Li \etal~\cite{li2018localizing} use convolutional neural networks and visualization methods to locate and quantify damage in a disaster images. Nalluru \etal~\cite{nalluru2019relevancy} combine semantic textual and image features to classify the relevancy of social media posts in emergency situations. 

Our framework focuses on combining images and text, yielding performance improvements on three disaster classification tasks. \\

\noindent\textbf{ Deep Multimodal Learning:}
In deep multimodal learning, neural networks are used to integrate the complementary information from multiple representations (modalities) of the same phenomena~\cite{wollmer2010context,ngiam2011multimodal,abavisani2018deep,chen2015multi,abavisani2019improving,perera2018in2i}.  In many applications, including image captioning \cite{bernardi2016automatic,rahman2019watch}, visual question answering \cite{antol2015vqa,fukui2016multimodal}, and text-image matching \cite{socher2013zero,frome2013devise,lee2018stacked}, combining image and text signals is of interest.  Thus many recent works study image-text fusion \cite{lu2019vilBERT,li2019visualbert,tan2019lxmert,su2019vl}.  

Existing multimodal learning frameworks applied to the crisis domain are relatively limited. Lan \etal~\cite{lan2014multimedia} combine early fusion and late fusion methods to incorporate their advantages, Ilyas \cite{ilyas2014microfilters} introduce a disaster classification system based on naive-bayes classifiers and support vector machines. Kelly \etal~\cite{kelly2017mining} introduce a system for real-time extraction of information from
text and image content in Twitter messages with exploiting the spatio-temporal metadata for filtering, visualizing, and monitoring flooding events. Mouzannar \etal~\cite{mouzannar2018damage} propose a multimodal deep learning framework to identify
damage related information on social media posts with texts, images, and video. 

In the application of crisis tweets categorization, one modality may contain uninformative or even misleading information. The attention module in our model passes information based on the confidence in the usefulness of different modalities. The more confident modality blocks weak or misleading features from the other modality through their cross-attention link. The partially blocked results of both modalities are later judged by a self-attention layer to decide which information should be passed to the next layer.  While our attention module is closely related to co-attention and self-attention mechanisms~\cite{vaswani2017attention,hessel2019case,liu2018learn,fukui2016multimodal,NIPS2017_6658,rahman2019watch}, unlike them, it does not need the input features to be homogeneous.   In contrast, self-attention and co-attention layers can be sensitive to heterogeneous inputs.  The details of the model are described in the next section.

%
%
%
%

\section{Methodology}

\begin{figure}[t!]
    \centering
    \includegraphics[width=0.5\textwidth]{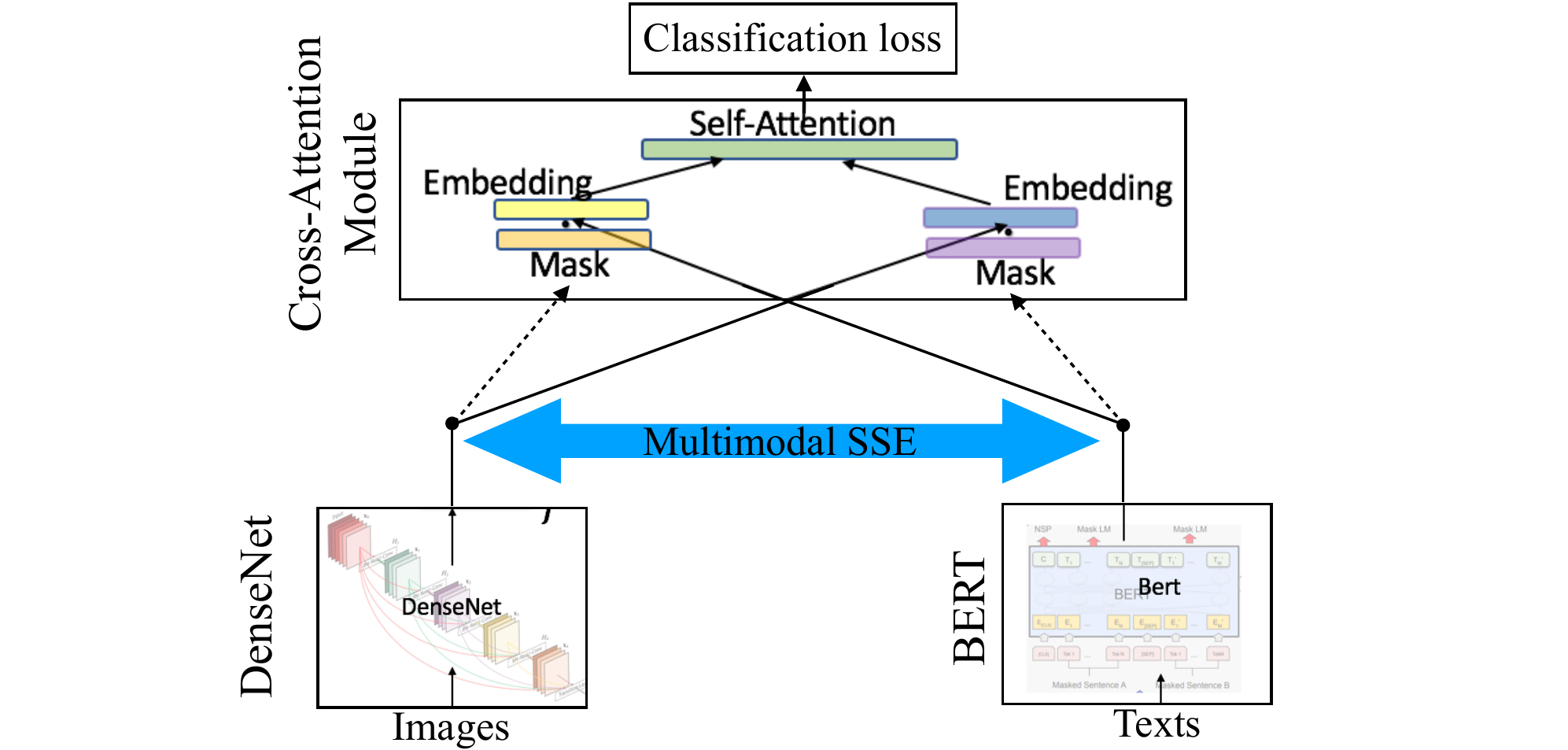}
    \label{fig:diagram}
\caption{Illustration of Our Framework. Embedding features are extracted from images and texts by DenseNet and BERT networks, respectively, and are integrated by the cross-attention module. In the training process, the embeddings of different samples are stochastically transitioned between each other to provide a robust regularization. }
\end{figure}

The architecture we propose is designed for classification problems that takes as input image-text pairs such as user generated tweets in social media, as illustrated in Figure~\ref{fig:diagram}, where the DenseNet and BERT graphs are from ~\cite{huang2017densely} and \cite{devlin2018BERT}. Our methodology consists of 4 parts: the first two parts extract feature maps from the image and extract embeddings from the text, respectively; the third part comprises our cross-attention approach to fuse projected image and text embeddings; and the fourth part uses Stochastic Shared Embeddings (SSE) \cite{wu2019stochastic} as our regularization technique to prevent over-fitting and deal with training data with inconsistent labels for image and text pairs.
 
We describe each module in the sub-sections that follow. 

\subsection{Image Model for Feature Map Extraction:}
We extract feature maps from images using Convolutional Neural Networks (CNNs). In our model we select DenseNet ~\cite{huang2017densely}, which reduces module sizes and increases connections between layers to address parameter redundancy and improve accuracy (other approaches, such as EfficientNet \cite{tan2019efficientnet} could also be used, but DenseNet is efficient and commonly used for this task). 

 For each image $v_i$, we therefore have: 
\begin{equation}
    f_i = \textbf{DenseNet}(v_i),
\end{equation} where $v_i$ is the input image, $f_i \in \mathbb{R}^{D_f}$ is the vectorized form of a deep feature map in the DenseNet with dimension $D_f = W \times H \times C$, where $W, H, C$ are the feature map's height, width and number of channels respectively.

\subsection{Text Model for Embedding Extraction:}
Full-network pre-training \cite{radford2018improving, devlin2018BERT} has led to a series of breakthroughs in language representation learning.  Specifically, deep-bidirectional Transformer models such as BERT \cite{devlin2018BERT} and its variants \cite{yang2019xlnet, anonymous2020alBERT} have achieved state-of-the-art results on various natural language processing tasks by leveraging close and next-sentence prediction tasks as weakly-supervised pre-training.

Therefore, we use BERT as our core model for extracting embeddings from text (variants such as XLNET~\cite{yang2019xlnet} and ALBERT~\cite{anonymous2020alBERT} could also be used).
We use the BERT model pre-trained on Wiki and Books data\cite{data} on crisis-related tweets $t_i$'s. For each text input $t_i$, we have
\begin{equation}
    e_i = \textbf{BERT}(t_i),
\end{equation} where $t_i$ is a sequence of word-piece tokens and $e_i \in \mathbb{R}^{756}$ is the sentence embedding. Similar to the BERT paper \cite{devlin2018BERT}, we take the embedding associated with [CLS] to represent the whole sentence. 

In the next subsection we detail how DenseNet and BERT are fused.

\subsection{Cross-attention module for avoiding negative knowledge in fusion:}\label{subsec:fuse}

After we obtain the image feature map $f_i$ (DenseNet) and the sentence embedding $e_i$ (BERT), we use a new cross-attention mechanism to fuse the  information they represent.   In many text-vision tasks, the input pair can contain noise. In particular, in classification of tweets, one modality may contain non-informative or even misleading information.  In such a case, negative information transfer can occur.  Our model can mitigate the effects of one modality over another on a case by case basis.

To address this issue, in our cross-attention module, we use a combination of cross-attention layers and a self-attention layer. 
In this module, each modality can block the features of the other modality based on its confidence in the usefulness of its input. This happens with the cross-attention layer. The result of partially blocked features from both modalities is later fed to a self-attention layer to decide which information should be passed to the next layer. 

The self-attention layer exploits a fully-connected layer to project the image feature map into a fixed dimensionality $K$ (we use $K = 100$), and similarly project the sentence embedding so that:
\begin{align}\label{eq:project}
 \nonumber   \tilde{f}_i &= F(W_v^T f_i + b_v), \\
    \tilde{e}_i &= F(W_e^T e_i + b_e),
\end{align} where $F$ represents an activation function such as ReLU (used in our experiments) and both $\tilde{f}_i$ and $\tilde{e}_i$ are of dimension $K = 100$.

In the case of misleading information in one modality, without an attention mechanism (such as co-attention \cite{lu2019vilBERT}), the resulting $\tilde{f}_i$ and $\tilde{e}_i$ cannot be easily combined without hurting performance. Here, we propose a new attention mechanism called cross-attention (Figure \ref{fig:diagram}), which differs from standard co-attention mechanisms: the attention mask $\alpha_{v_i}$ for the image is completely dependent on the text embedding $e_i$, while the attention mask $\alpha_{e_i}$ for the text is completely dependent on the image embedding $f_i$. Mathematically, this can be expressed as follows:
\begin{align}\label{eq:cross-attnetion}
 \nonumber   \alpha_{v_i} &= \sigma({W'_v}^T f_i + b'_v), \\
    \alpha_{e_i} &= \sigma({W'_e}^T e_i + b'_e),
\end{align} where $\sigma$ is the Sigmoid function.
Co-attention, in contrast, can be expressed as follows:
\begin{align}\label{eq:co-attnetion}
 \nonumber   \alpha_{v_i} &= \sigma({W'_v}^T [f_i | e_i] + b'_v), \\
    \alpha_{e_i} &= \sigma({W'_e}^T [f_i | e_i] + b'_e),
\end{align} where $|$ means concatenation.

After we have the attention masks $\alpha_{v_i}, \alpha_{e_i}$ for image and text, respectively, we can augment the projected image and text embeddings $\tilde{f}_i, \tilde{e}_i$ with $\alpha_{v_i} \cdot  \tilde{f}_i$ and $\alpha_{e_i} \cdot  \tilde{e}_i$ before performing concatenation or adding. In our experiments, we use concatenation but obtained similar performance using addition.

The last step of this module takes the concatenated embedding which jointly represents the image and text tuple in and feeds into the two-layer fully-connected networks. We add self-attention in the fully-connected networks and use the standard softmax cross-entropy loss for the classification.

In Section~\ref{sec:exp}, we show that the combination of cross-attention layers and the self-attention layer on their concatenation works better than co-attention and self-attention mechanisms for the tasks we address in this paper. 

\subsection{SSE for Better Regularization}
Due to unforeseeable and unpredictable nature of disasters, and also because they require fast processing and reaction, one often has to deal with limited annotations for user-generated content during crises. Using regularization techniques to mitigate this issue becomes especially important.   In this section, we extend Stochastic Shared Embeddings (SSE) technique \cite{wu2019stochastic} to its multimodal version for taking the full advantage from the annotated data by 1) generating new artificial multimodal pairs. 2) also including the annotated data with inconsistent labels for text and image in the training process.  

SSE-Graph \cite{wu2019stochastic}, a variation of SSE, is a data-driven approach for regularizing embedding layers which uses a knowledge graph to stochastically make transitions between embeddings of different samples during the stochastic gradient descent (SGD). That means, during the training, based on a knowledge graph, there is a chance that embeddings of different samples being swapped.   We use the text and image labels to construct knowledge graphs that can be used to create stochastic multimodal training samples with consistent labels for both the image and text.  

We treat feature maps of images as embeddings and use class labels to construct knowledge graphs. The feature maps of two images are connected by an edge in the graph, if and only if they belong to the same class (e.g. they are both labeled ``affected individuals'').  We follow the same procedure for text embeddings and construct a knowledge graph for text embeddings as well. Finally, we connect the nodes associated with the knowledge graph of image feature maps with an edge to nodes in text's knowledge graph if and only if they belong to the same class.

Let $\Phi^v$ and $\Phi^t$ be sets of parameters. We define the transition probability  $p(i^{v}, j^{v}|\Phi^v)$ as probability of transition from $i^{v}$ to $j^{v}$, where $i^{v}$ and $j^{v}$ are nodes in the image knowledge graph that correspond to image features $f_i$ and $f_j$. Similarly, we define $p(i^{t}, k^{t}|\Phi^t)$ as probability of transition from $i^{t}$ to $k^{t}$ (nodes corresponding to text embeddings $e_i$ and $e_k$, respectively). 

Taking image feature maps as an example, if $i^v$ is connected to $j^v$ but not connected to $l^v$ in the knowledge graph, one simple and effective way to generate more multimodal pairs is to use a random walk (with random restart and self-loops) on the knowledge graph.  Since we are more interested in transitions within embeddings of consistent labels, in each transition probability, we set the ratio of $p(i^v, j^v|\Phi^v)$ and $p(i^v, l^v|\Phi^v)$ to be a constant greater than $1$. In more formal notation, we have
\begin{equation}\label{eq:transition1}
   i^v \sim j^v, i^v \not\sim l^v \longrightarrow  p(i^v, j^v|\Phi^v)/p(i^v, l^v|\Phi^v) = \rho^v, 
\end{equation} where $\rho^v$ is a tuning parameter and $\rho^v > 1$ , and $\sim$ and $\not\sim$ denote connected and not connected nodes in the knowledge graph. We also have:
\begin{equation}\label{eq:transition2}
    p(i^v, i^v|\Phi) = 1 - p_0^v,
\end{equation} 
where $p_0^v$ is called the {\it SSE probability} for image features. 

We similarly define $\rho^t$ and $p_0^t$ in $\Phi^t = \{\rho^t,p_0^t\}$ for text embeddings. Note that $\rho^t$ is defined with respect to the image features' label. That is
\begin{equation}\label{eq:transition1}
   i^v \sim j^t, i^v \not\sim l^t \longrightarrow  p(i^t, j^t|\Phi^t)/p(i^t, l^t|\Phi^t) = \rho^t. 
\end{equation}

Both $\Phi^v$ and $\Phi^t$ parameters sets are treated as tuning hyper-parameters in experiments and can be tuned fairly easily. With Eq. \eqref{eq:transition1}, Eq. \eqref{eq:transition2} and $\sum_{k^v} p(j^v, k^v|\Phi^v), \sum_{k^t} p(j^t, k^t|\Phi^t) = 1$, we can derive transition probabilities between any two sets of feature maps in images and texts to fill out the transition probability table. 

With the right parameter selection, each multimodal pair in the training can be transitioned to many more multimodal pairs that are highly likely to have consistent labels for the image and text pairs which can mitigate both the issues of limited number of training samples and inconsistency in the annotations of image-text pairs.

%
%
%
%


\section{Experimental Setup}\label{sec:exp}

The image-text classification problem we consider can be formulated as follows: we have as input $(v_1, t_1), \dots, (v_i, t_i), \dots, (v_n, t_n)$, where $n$ is the number of training tuples and the $i$-th tuple consists of both image $v_i$ and text $t_i$. The respective labels for $v_i$ and $t_i$'s are also given in training data. Our goal is to predict the correct label for any unseen $(v, t)$ pair. To simplify the evaluation, we assume there is only one correct label associated with the unseen $(v, t)$ pairs. As a result, this paper targets a multi-class classification problem instead of a multi-label problem.

\subsection{Dataset}
There are very few crisis datasets, and to the best of our knowledge there is only one \textit{multimodal} crisis dataset, CrisisMMD~\cite{alam2018crisismmd}. It consists of annotated image-tweet pairs where images and tweets are independently labeled as described below.  We use this dataset for our experiments.  
The dataset was collected using event-specific keywords and hashtags during seven natural disasters in 2017: Hurricane Irma, Hurricane Harvey, Hurricane Maria, the Mexico earthquake, California wildfires, Iraq-Iran earthquakes, and Sri Lanka floods.   The corpus is comprised of three types of manual annotations: \\

    \noindent \textbf{Task 1:} Informative vs. Not Informative: whether a given tweet text or image is useful for humanitarian aid purposes, defined as providing assistance to people in need.\\ 
    
    \noindent \textbf{Task 2:} Humanitarian Categories: given an image, or tweet, or both, categorize it into one of the five following categories:
    \begin{itemize}
        \item Infrastructure and utility damage
        \item Vehicle damage
        \item Rescue, volunteering, or donation efforts
        \item Affected individuals (injury, dead, missing, found, etc.)
        \item Other relevant information
    \end{itemize} 
    Note that we merge the data that are labeled as \emph{injured or dead people} and  \emph{missing or found people} in the CrisisMMD with those that are labeled as \emph{affected individuals} and view all of them as one class of data.\\

    \noindent \textbf{Task 3:} Damage Severity: assess the severity of damage reported in a tweet image and classify it into Severe, Mild, and Little/None. \\
    
\noindent It is important to note that while the annotations for the last task are only on images. Our experiments reveal that using tweet texts along with the images can boost performance. In addition, our paper is the first one to perform all three tasks on this dataset (text-only, image-only, combined).  

\subsection{Settings}
Images and text from tweets in this dataset were annotated independently. Thus, in many cases, images and text in the same pairs may not share the same labels for either Task 1 or Task 2 (labels for Task 3 were only created by annotating the images). 
Given the different evaluation conditions, we carry out three evaluation settings for the sake of being comprehensive in our model assessment but also to establish best practices for the community: 
\emph{Setting A}: we exclude the image-text pairs with differing labels for image and text; \emph{Setting B}: we include the image-text pairs with different labels in the training set but keep the test set the same as in A.  

In addition, we introduce \emph{Setting C} to mimic a realistic crisis tweet classification task where we only train on events that have transpired before the event(s) in the test set. 

Table~\ref{tbl:sss} shows the number of samples in each set for different setting and tasks.\\

\begin{table}[t]
\caption{Number of samples in different splits of our settings.}
\resizebox{\columnwidth}{!}{
\begin{tabular}{lccc}
\toprule
     Setting           &  \# of Training samples       & \# of Dev samples   & \# of Test samples\\ \hline
Setting A    &             &     &     \\ \hline
\hfill Task1:    &         7876      &  553    &  2821   \\ 
\hfill Task2:    &          1352     &   540   &    1467 \\ 
\hfill Task3:    &          2590     &   340   &   358  \\ \hline
Setting B   &            &      &     \\ \hline
\hfill Task1:    &      12680         &  553    &   2821  \\
\hfill Task2:    &        5433       &  540    &  1467   \\ \hline
Setting C  &            &      &     \\ \hline
\hfill Experiment 1:  &       174    &   -   &   217  \\ 
\hfill Experiment 2:     &       4037        &  -    & 217    \\ 
\hfill Experiment 3:     &      4761         &   -   &  217   \\     \bottomrule
\end{tabular}
}
\label{tbl:sss}
\end{table}

\noindent\textbf{Setting A:} 
In this setting our train and test data is sampled from tweets in which the text and image pairs have the same label. That is:
\begin{equation}\label{eq:assumption}
    \mathcal{C}(v_i) = \mathcal{C}(t_i),
\end{equation}
where $\mathcal{C}(x)$ denotes the class of data point $x$. This results in a small, yet potentially more reliable training set. We mix the data from all seven crisis events and split the data into training, dev and test sets. \\

\begin{table*}
  \caption{Setting A: Informativeness Task, Humanitarian Categorization Task  and Damage Severity Task Evaluations.}
  \label{tb:s1}
  \centering
 \resizebox{\textwidth}{!}{
  \begin{tabular}{cccccccccc}
    \toprule
    & \multicolumn{3}{c}{Informativeness Task}   & \multicolumn{3}{c}{Humanitarian Categorization Task}&\multicolumn{3}{c}{Damage Severity Task}              \\
    \cmidrule(r){2-4} \cmidrule(r){5-7}  \cmidrule(r){8-10}  
    Model     &  Acc & Macro F1 & Weighted F1  & Acc & Macro F1 & Weighted F1& Acc & Macro F1 & Weighted F1\\
    \midrule
    DenseNet \cite{huang2017densely} &  81.57 &79.12 & 81.22  &83.44 &60.45 & 86.96& 62.85 &52.34 & 66.10 \\
    BERT \cite{devlin2018BERT} &   84.90 &81.19 & 83.30& 86.09 &66.83 & 87.83 & 68.16 &45.04 & 61.09 \\
     \midrule
   
    Compact Bilinear Pooling\cite{fukui2016multimodal}  &  88.12 &86.18 & 87.61&89.30 & 67.18 & 90.33& 66.48 & \bf{61.03} & \bf{70.58} \\
     Compact Bilinear Gated Pooling \cite{kiela2018efficient} & 88.76 & 87.50 & 88.80&85.34 &65.95 & 89.42  & 68.72 &51.46 & 65.34\\
      MMBT~\cite{kiela2019supervised}& 82.48&81.27&82.15&85.82& 64.78& 88.66&65.36&52.12&69.34\\
       \midrule
      Score Fusion  & 88.16 & 83.46 & 85.26&86.98 &54.01 & 88.96 & 71.23 &53.48 & 66.26 \\
    Feature Fusion &     87.56 &85.20 & 86.55 &89.17 &67.28 & 91.40& 67.60 &40.62 & 56.47 \\
   
    \midrule
Attention Variant 1~\footnotesize{(Ours)} & 89.29 &85.68 & 87.04& 88.41 &64.60 & 90.71 & 71.51 &55.41 & 69.71 \\ 
Attention Variant 2~\footnotesize{(Ours)}  & 88.34 &86.12 & 87.42&   89.23 & 67.63 & 91.56& 63.13 &58.03 & 69.39 \\
Attention Variant 3~\footnotesize{(Ours)}  & 88.20 &86.22 & 87.47& 87.18 &64.67 & 90.24 & 68.99 &57.42 & 69.16 \\
\midrule
SSE-Cross-BERT-DenseNet~\footnotesize{(Ours)} &  \bf{89.33} & \bf{88.09} & \bf{89.35} & \bf{91.14} &\bf{68.41} & \bf{91.82} & \bf{72.65} & 59.76 & 70.41  \\
    \bottomrule
  \end{tabular}
 }
\end{table*}

\noindent\textbf{Setting B}: We relax the assumption in Equation~\ref{eq:assumption} and allow in training:
\begin{equation}\label{eq:assumption2}
    \mathcal{C}(v_i) \neq \mathcal{C}(t_i),
\end{equation}

As the training set of this setting contains samples with inconsistent labels for image and text,  multimodal fusion methods such as late feature fusion cannot deal with the training data. Our method, on the other hand, with the use of the proposed multimodal SSE, can transition the training instance with in consistent labels to a new training pair with consistent labels. We do this by manually setting $p^t_0=1$ for the training cases with inconsistent image-text labels (i.e. all the text samples are transitioned). Since unimodal models only receive one of the modalities, it is also possible to train them separately on images and texts and use an average of their prediction in the testing stage (also known as score level fusion).

However, we maintain the assumption of Eq. \eqref{eq:assumption} for the test data. This helps to directly compare the two settings with the same test samples. In fact, in practice, the data is most valuable when the class labels match for both image and text. The rationale is that detecting an event is more valuable to crisis managers than the categorization of different parts of that event. Our dev and test sets for this setting are similar to the previous setting. However, the training set contains a larger number of samples where their image-text pairs are not necessarily labeled as the same class. \\

\noindent\textbf{Setting C:} This setting is closest to the real-world scenario where we analyze the new event of a crisis with a model trained on previous crisis events.  
First, we require the training and test sets to be from crisis events of a different nature (i.e., wildfire vs. flood). Second, we maintain the temporal component and only train on events that have happened before the tweets of the testing set.  Since collecting annotated data on an urgent ongoing-event is not possible, and also because an event of crisis may do not have a similar annotated event in the past, these two restrictions often simulate a real-world scenario.  For the experiments of this setting, there is no dev set. Instead, we use a random portion of the training data to tune the hyper-parameters.

We test on the tweets that are related to the California Wildfire (Oct 10 - 27, 2017), and train on the following three sets:

    \begin{compactenum}
         \item Sri Lanka Floods tweets (\textit{May 31- Jul. 3, 2017})
        \item Sri Lanka Floods, and Hurricane Harvey and Hurricane Irma tweets (\textit{May 31- Sept. 21 , 2017})
        \item Sri Lanka Floods, Hurricanes Harvey and Irma and Mexico Earthquakes  (\textit{May 31 - Oct. 5, 2017}). 
    \end{compactenum}

Similar to setting B, for the test set (i.e. California Wildfire) we only consider the samples with consistent labels for image and text, but for the training sets, we use all the available samples.

\begin{table}[tb]
  \caption{Setting B: Informativeness Task and Humanitarian Categorization Task Evaluations}
  \label{tb:tasks-s2}
  \centering
 \resizebox{\columnwidth}{!}{
  \begin{tabular}{ccccccc}
    \toprule
    & \multicolumn{3}{c}{Informativeness Task}   & \multicolumn{3}{c}{Humanitarian Categorization Task}  \\
    \cmidrule(r){2-4} \cmidrule(r){5-7}   
     Model     &  Accuracy & Macro F1   & Weighted F1   &  Accuracy & Macro F1   & Weighted F1    \\
    \midrule
    DenseNet \cite{huang2017densely} &  83.36 & 80.95 & 82.95 &   82.89 &  66.68 &83.13 \\
    BERT \cite{devlin2018BERT} &  86.26 & 84.44 & 86.01  &   87.73 &  83.72 & 87.57  \\
    Score Fusion &  87.03 & 85.19 & 86.90 & 91.41 &  83.26 &  91.36 \\
\midrule
SSE-Cross-BERT-DenseNet~\footnotesize{(Ours)}  & \textbf{90.05} & \textbf{88.88} & \textbf{89.90}  &  \textbf{93.46} & \textbf{84.16} & \textbf{93.35} \\ 
\midrule
Best from Table \ref{tb:s1} &   89.33 & 88.09 & 89.35  &91.48 & 67.87& 91.34 \\ 
    \bottomrule
  \end{tabular}
 }
\end{table}

\begin{table*}[tb]
 \caption{Comparing our proposed method with baselines for Humanitarian Categorization Task in Setting 3. We fix the last occurred crisis namely `California wildfires' as test data and vary the training data which is specified in the columns.}
  \label{tb:tasks-s3}
  \centering
 \resizebox{\textwidth}{!}{
  \begin{tabular}{cccccccccc}
    \toprule
    & \multicolumn{3}{c}{Sri Lanka Floods}   & \multicolumn{3}{c}{Sri Lanka Floods + Hurricanes Harvey \& Irma}   & \multicolumn{3}{c}{Sri Lanka Floods + Hurricanes Harvey \& Irma + Mexico earthquake}\\
    \cmidrule(r){2-4} \cmidrule(r){5-7}   \cmidrule(r){8-10} 
    Model     &  Accuracy & Macro F1   & Weighted F1   &  Accuracy & Macro F1   & Weighted F1  &  Accuracy & Macro F1   & Weighted F1  \\
    \midrule
    DenseNet \cite{huang2017densely} &  55.71 &35.77 & 56.85 &  70.32 & 52.23 & 68.55 &70.32 & 44.80 & 68.79 \\
    BERT \cite{devlin2018BERT} & 31.96 & 20.90 & 27.21  &  73.97 & 53.90 & 73.51 & 74.43 & 56.98 & 74.21\\
    Score Fusion  &   56.62 & 36.77 & 57.96 & 81.74 &  56.54 & 81.03 & 81.28 &  55.90 & 80.54\\
\midrule
SSE-Cross-BERT-DenseNet~\footnotesize{(Ours)}  &  \textbf{62.56} & \textbf{39.82} & \textbf{62.08}  & \textbf{84.02} & \textbf{63.12} & \textbf{83.55} & \textbf{86.30} & \textbf{65.55} & \textbf{85.93} \\ 
    \bottomrule
  \end{tabular}
  
 }
  
\end{table*}

\begin{table}
  \caption{Ablation Study of our proposed method for Humanitarian Categorization Task in Setting A.}
  \label{tb:ablation}
  \centering
 \resizebox{\columnwidth}{!}{
  \begin{tabular}{cccc}
    \toprule
    & \multicolumn{3}{c}{Test Set}              \\
    \cmidrule(r){2-4} 
    Model     &  Accuracy & Macro F1 & Weighted F1  \\
    \midrule
SSE-Cross-BERT-DenseNet~\footnotesize{(Ours)} &   \textbf{91.14} & \textbf{68.41} & \textbf{91.82} \\
\midrule
$-$ Self-Attention &89.23 & 56.50 & 87.70  \\
$-$ Cross-Attention &  88.48 & 56.38 & 87.10  \\
$-$ Cross-Attention $+$ Co-Attention &  88.41 & 64.60 & 90.71  \\
$-$ Cross-Attention $+$ Self-Attention &86.30 & 58.33 & 85.27  \\ 
$-$ Dropout & 83.37 & 54.83 &82.46  \\
$-$ SSE &  88.41 &64.60 & 90.71   \\
$-$ SSE $+$ Shuffling Within Class & 88.68 & 62.91 &88.33  \\
$-$ SSE $+$ Mix-up \cite{zhang2018mixup} &   89.16 & 54.63 & 87.37   \\
    \bottomrule
  \end{tabular}
 }
\end{table}

\subsection{Baselines}
We compare our method against several state-of-the-art methods for text and/or image classification.  
There are a number of categories of baseline methods we compare against.
In the first category, we compare to DenseNet and BERT, which are of the most comonnly used unimodal classification networks for images and texts respectively. We use Wikipedia pre-trained BERT and pre-trained DenseNet on ImageNet~\cite{deng2009ImageNet}, and fine-tune them on the training sets. 

The second category of baseline methods include several recently proposed multimodal fusion methods for classification: 
\begin{itemize}
    \item Compact Bilinear Pooling \cite{fukui2016multimodal}: multimodal compact bilinear pooling is a fusion technique first used in visual question answering task but can be easily modified to perform standard classification task.
    \item Compact Bilinear Gated Pooling \cite{kiela2018efficient}: this fusion method is an adaptation of the compact bilinear pooling method where an extra attention gate is added on top the compact bilinear pooling module.
    \item MMBT~\cite{kiela2019supervised}: recently proposed supervised multimodal bitransformers model for classifying images and text.
\end{itemize}

The third category is the score level \emph{Score Fusion} and late feature fusion \emph{Feature Fusion} of DenseNet and BERT networks. 
Score level fusion is one of the most common fusion techniques.  It averages the predictions of separate networks trained on the different modalities.  Feature Fusion is one of the most effective methods for integrating two modalities~\cite{ramachandram2017deep}.  It concatenates deep layers from modality networks to predict a shared output.
We also provide three variations of our attention modules and report their performance:
 The first variant is to replace cross-attention of Eq.~\eqref{eq:cross-attnetion} with co-attention of Eq.~\eqref{eq:co-attnetion}; the second variant is to remove self-attention; the third variant is to change the cross-attention with self-attention modules.

We compare our model, SSE-Cross-BERT-DenseNet, to the baseline models above.

\subsection{Evaluation Metrics} We evaluate the models in this paper using classification accuracy,\footnote{In the settings that our experiments are defined classification accuracy is equivalent to Micro F1-score.} Macro F1-score and weighted F1-score.  Note that while in the event of a crisis, the number of samples from different categories often significantly varies, it is important to detect all of them. F1-score and weighted F1-score take both false positives and false negatives into account, and therefore, along with accuracy as an intuitive measure, are proper evaluation metrics for our datasets.

\subsection{Training Details}
We use pre-trained DenseNet and BERT as our image and text backbone networks, and fine-tune them separately on text-only and image-only training samples. The details of their implementations can be found in \cite{huang2017densely} and  \cite{devlin2018BERT}, respectively.  We do not freeze the pre-trained weights and train all the layers for both the backbone networks. 

We use the standard SGD optimizer.   We start with the base learning rate of $2\times10^{-3}$ with a $10\times$ reduction when the dev loss is saturated.  We use a batch size of $32$. The models were implemented in Keras and Tensorflow-1.4~\cite{abadi2016tensorflow}. In all the applicable experiments, we select hyper-parameters with cross-validation on the accuracy of dev set. For the experiments in Setting 3 that we do not have an evaluation set, we tune hyper-parameters on $15\%$ of the training samples.  We select $\rho^v,\rho^t$ and $p_0^v,p_0^t$ respectively in the range of $\rho^v,\rho^t \in [ 10, 20000]$ and $p_0^v,p_0^t \in [0, 1]$.

We employ the following data augmentations on the images during the training stage. Images are resized such that the smallest side is 228 pixels, and then randomly cropped with a $224\times224$ patch.   In addition, we produce more images by randomly flipping the resulting image horizontally.  

For tweet normalization, we remove double spaces and lower case all characters. In addition, we replace any hyperlink in the tweet with the sentinel word ``\texttt{link}''.

\section{Experimental Results}
\subsection{Setting A: Excluding The Training Pairs with Inconsistent Labels}
As shown in Table~\ref{tb:s1}, our proposed framework, SSE-Cross-BERT-DenseNet, easily outperforms the standalone DenseNet and BERT models. 
Compared with baseline methods Compact Bilinear Pooling \cite{fukui2016multimodal}, Compact Bilinear Gated Pooling \cite{kiela2018efficient}, and MMBT \cite{kiela2019supervised}, our proposed cross-attention fusion method does enjoy an edge over previous known fusion methods, including the standard score fusion and feature fusion.  This edge holds true across Settings A, B and C. In section~\ref{sec:ablation}, we conduct an ablation study to investigate which components (SSE, cross-attention, and self-attention) have the most impact on model performance.

One important observation we find across the three tasks is that despite the fact that accuracy percentages are reasonably good for simple feature fusion method, the macro F1 scores improve much more once we add attention mechanisms.

\subsection{Setting B: Including The Training Pairs with Inconsistent Labels}
In this setting, we investigate whether our models can perform better if we can make use of more labelled data for un-matched images and texts.  Note that this involves training on noisier data than the prior setting.  
In Table~\ref{tb:tasks-s2}, our proposed framework SSE-Cross-BERT-DenseNet beats the best results from Setting A for both the Informativeness Task (89.90 to 89.35 Weighted F1) and the Humanitarian Categorization Task (93.35 to 91.34).  The gap between our method versus standalone BERT and DenseNet also widens.   

Note that the test sets are the same for setting A and setting B while only the training data differs.

\subsection{Setting C: Temporal}
This setting is designed to resemble a realistic scenario where the available data is (1) only from the past (i.e. the train / test sets are split in the order they occurred in the real world). (2) train and test sets are not from the same crisis. We find  that our proposed model consistently performs better than standalone image and text models (see Table~\ref{tb:tasks-s3}). Additionally, performance increases for all models, including ours, with the inclusion of more crisis data to train on.   This emphasizes the importance of collecting and labelling more crisis data even if there is no guarantee that the crises we collected data from will be similar to a future one. In the experiments, training crises contain floods, hurricanes and earthquakes but the test crisis is fixed at wildfires.

\subsection{Ablation Study}\label{sec:ablation}
In our ablation study, we examine each component of the model in Figure~\ref{fig:diagram}: namely self-attention on concatenated embedding, cross-attention on fusing image feature map \& sentence embedding, dropout and SSE regularization. All the experiments in this section are conducted in Setting A. First, we find self-attention plays an important role on the final performance, accuracy drops to 89.23 from 91.14 if self-attention is removed. Second, the choice of cross-attention over co-attention and self-attention is well justified: we see the accuracy performance drops to around 88 by replacing the cross-attention. Third, dropout regularization \cite{srivastava2014dropout} plays an important role in regularizing the hidden units: if we remove dropout completely, performance suffers a large drop from 91.14 to 83.37. Fourthly, we justify the usage of SSE \cite{wu2019stochastic} over the choice of Mixup \cite{zhang2018mixup} or within-class shuffling data augmentation. SSE performs better than mixup in terms of accuracy 91.14\% versus 89.16\%, and even much better in terms of F1 scores, 68.41 versus 54.63 for macro F1 score and 91.82 versus 87.37 for weighted F1 score.

\section{Conclusions and Future Work}
In this paper, we presented a novel multimodal framework for fusing image and textual inputs.  We introduced a new cross attention module that can filter not-informative or misleading information from modalities and only fuse the useful information.  We also presented a multimodal version of Stochastic Shared Embeddings (SSE) to regularize the training process and deal with limited training data.   We evaluate this approach on three crisis tasks involving social media posts with images and text captions.  We show that our approach not only outperforms image-only and text-only approaches which have been the mainstay in the field, but also other multimodal combination approaches.  

For future work we plan to test how our approach generalizes to other multimodal problems such as sarcasm detection in social media posts \cite{castro-etal-2019-towards, Schifanella_2016}, as well as experiment with different image and text feature extractors.  Given that the CrisisMMD corpus is the only dataset available for this task and it is limited in size, we also aim to construct a larger set, which is a major effort.

\newpage

{\small
\bibliographystyle{ieee_fullname}
\bibliography{egbib}
}

\clearpage

 \section*{Appendix: Setting D Multi-Label Multi-class Categorization}

In previous experiments of this paper, we followed prior research in crisis event categorization and viewed the task as a multi-class single-label task. In this section, we provide three simple modifications to our model for extending it to a multi-label multi-class classifier. 

\begin{figure}[t]    \label{fig:settings}
    \centering
    \includegraphics[width=.49\textwidth]{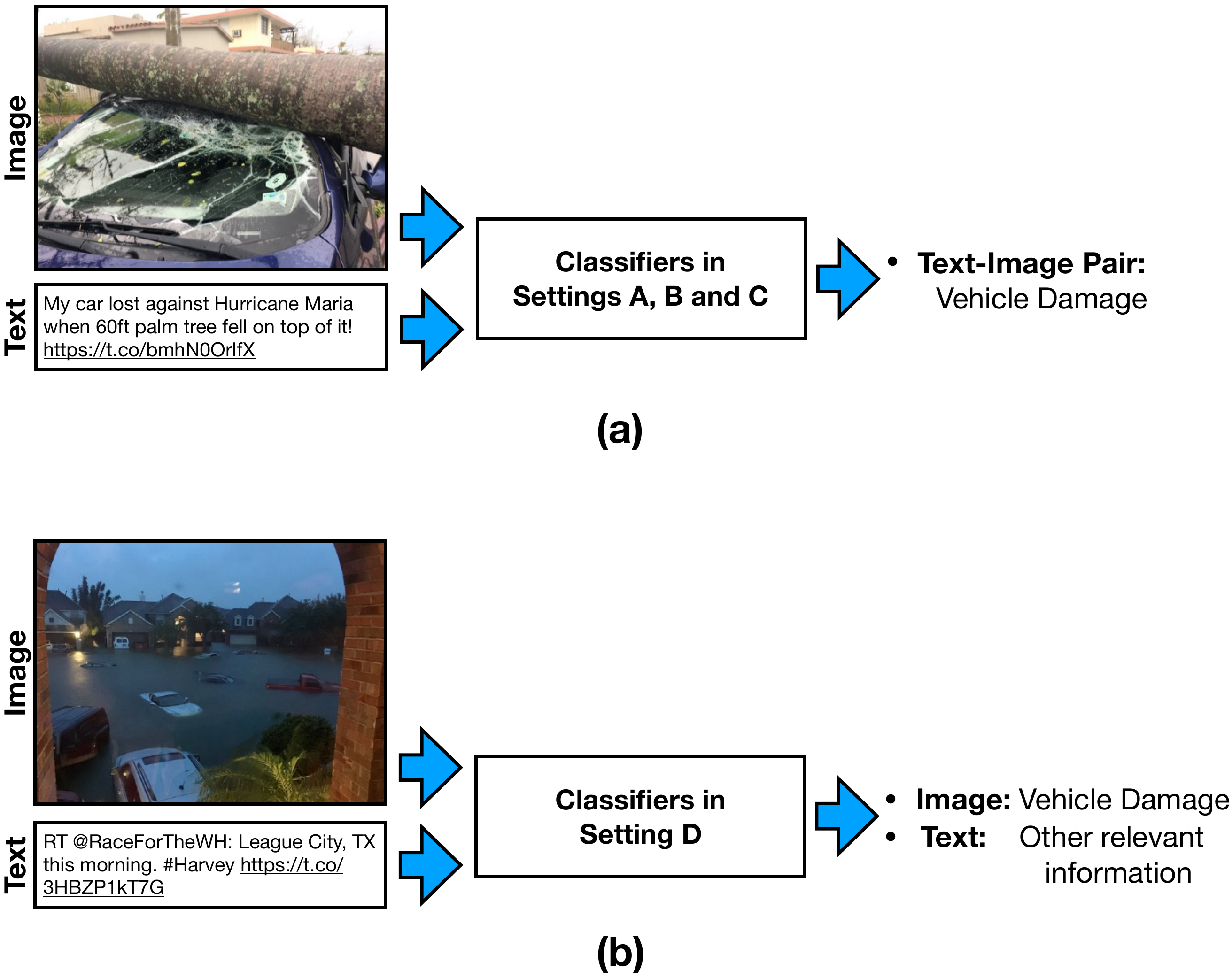}
    \caption{The behavior of our classifiers in different settings. (a) Our classifiers in settings A, B, and C view the task as a multi-class single-label task. (b) Our classifiers in setting D view the task as a multi-class multi-label task. }
\end{figure}

In a multimodal single-label classification system, representations of different modalities are often fused to construct a joint representation from which a common label is reasoned for the multimodal-pair. Our classifiers in settings A, B, and C are multimodal multi-class single-label models. However, in setting D, we are interested in using both image and text information to predict separate labels for them. Figure~\ref{fig:settings} (a) and (b) show examples of these settings.  

In Figure~\ref{fig:settings}~(a), the multimodal pair, including image and text are both labeled as \emph{Vehicle Damage}. On the contrary, in Figure~\ref{fig:settings}~(b), while the image shows damaged vehicles, the text-only contains information about the location of the event and therefore does not fall in the~\emph{Vehicle Damage} category. In setting D, we want to use the information in both image and text to classify the image of this example into the~\emph{Vehicle Damage} class and the text into the~\emph{Other Relevant Information} class.\\

\noindent\textbf{Cross-Attention:} A straightforward way to capture these properties is by attaching two classifier heads to the output of the cross-attention module in our proposed model. We refer to this version as \emph{Cross-Attention} classifier.   \\

\noindent\textbf{Self-Attention:}  The cross-attention mechanism in Eq. (4) uses text embeddings (image feature maps) to block misleading information from image feature maps (text embeddings).  However, in setting D, since image and text may have different labels, they both can be informative but contain different information. Thus, we replace this module by separate self-attention blocks~\cite{fukui2016multimodal,NIPS2017_6658} in each modality. That is, we still filter the uninformative features, but we do that based on the information in the modality itself. \\

\noindent\textbf{Self-Cross-Attention:}  In the \emph{Self-Attention} extension, the features of different modalities do not interact directly with each other. With a few modifications to the self-attention extension and combining it with our cross-attention model, one can develop a version of our method that is specifically designed for multi-label multi-class classification tasks. We use a self-attention block to learn a mask that filters the uninformative features from the modalities. In the meantime, we invert this mask and use the invert mask to attend to the other modality for selecting useful features.     This way, not only do we develop modality-specific features, but we do so by exploiting useful information from both modalities. Let $\gamma_{v_i}$ and $\gamma_{t_i}$ be the self-attention masks that are calculated as:
\begin{align}\label{eq:self-attnetion}
 \nonumber   \gamma_{v_i} &= \sigma({W''_v}^T [f_i] + b''_v), \\
    \gamma_{e_i} &= \sigma({W''_e}^T [e_i] + b''_e)
\end{align}
From equation~\eqref{eq:self-attnetion}, we can calculate the inverse-masks by
\begin{align}\label{eq:self-cross-attnetion}
 \nonumber   \gamma^{\prime}_{v_i} &= 1- \gamma_{v_i} \\
     \gamma^{\prime}_{e_i} &=1- \gamma_{e_i}.
\end{align} 

After we have the attention masks and the inverse of them, we can calculate the augmented image features ${f''}_i$ and augmented text feature ${e''}_i$ as
\begin{align}\label{eq:self-cross-attnetion}
\nonumber {f''}_i &=  \gamma{v_i} \cdot  \tilde{f}_i + \gamma^{\prime}_{v_i} \cdot  \tilde{e}_i\\
 {e''}_i &=  \gamma{t_i} \cdot  \tilde{e}_i + \gamma^{\prime}_{t_i} \cdot  \tilde{f}_i
\end{align} 
where $\tilde{e}_i$ and $\tilde{f}_i$ are same as in Eq. (3) in the paper.  We feed  ${f''}_i$ and $ {e''}_i$ to classifier heads of images and texts, respectively.\\ 

\subsection{Experiments:}

We evaluate the multi-label extensions in Task 1. In this experiment, both training and test sets contain inconsistent labels.  That is in both training and testing we may have:

\begin{equation}\label{eq:assumption}
    \mathcal{C}(v_i) \neq \mathcal{C}(t_i),
\end{equation}

\begin{table}
  \caption{Setting D: Informativeness Evaluation}
  \label{tb:tasks-s4}
  \centering
 \resizebox{\columnwidth}{!}{
  \begin{tabular}{lccccc}
    \toprule
    Model     &&  Acc & Macro F1 & Weighted F1   \\
    \midrule
    DenseNet \cite{huang2017densely}&Images :&  78.30 &78.30 & 78.31 \\

    BERT \cite{devlin2018BERT}&Text :& 82.63 & 74.93 & 80.87 \\
         \midrule
\multirow{2}{*}{Feature Fusion} &Images :& 78.37 & 78.15 & 78.21 \\ 
 &Texts:& 83.63 & 79.01 & 83.22 \\ 
          \midrule
\multirow{2}{*}{Cross-Attention} &Images :& 77.17 &77.51 & 77.51 \\ 
 &Texts:&83.35 &\textbf{79.60} & \textbf{83.41} \\ 
         \midrule
\multirow{2}{*}{Self-Attention} &Images :& \textbf{82.56} & \textbf{82.54} & \textbf{82.56} \\ 
 &Texts:&\textbf{83.63} & 76.79 & 82.17 \\ 
          \midrule
\multirow{2}{*}{Self-Cross-Attention} &Images :& 81.64 & 81.51 & 81.55 \\ 
 &Texts:& 83.45 & 78.22 & 82.78 \\   
    \bottomrule
  \end{tabular}
 }
\end{table}

As the test set of this setting contains samples with inconsistent labels for image and text,  we set $0<p^t_0<1$ for the training cases so that we include inconsistent image-text labels in training as well. In particular, we use $\Phi_t=\{p^t_0:0.27,\rho_t:900\}$ and $\Phi_v=\{p^v_0:0.36,\rho_v:900\}$. Benchmarks for this setting include unimodal models as well as a version of the feature fusion model with two classification heads.

We evaluate our method on Task 1. We keep the ratio between the number of samples in train and test sets similar to setting B in Table 2. However, we randomly sample with relaxing the Eq. (9) assumption of the paper for both the train and test sets. 

In Table~\ref{tb:tasks-s4}, the result of different methods are compared in terms of Accuracy, Macro-F1, and Weighted F1. By comparing unimodal DenseNet and BERT results with Table 4, we observe that the test set in setting D, with inconsistent labels for images and texts, is more challenging than the test set in previous settings. As can be seen,  most methods have an advantage over unimodal DenseNet and BERT. The Cross-Attention method provides better results for text, and Self-Attention method provides better results for images. The Self-Cross-Attention, on average, provides comparable results to the Self-Attention and Cross-Attention methods for both the modalities. Note that in all three attention methods, the multimodal-SSE technique has been used, which provides additional training data (with both consistent and inconsistent labels).
\end{document}